\documentclass[letterpaper, 10 pt, conference]{ieeeconf}  

\IEEEoverridecommandlockouts                              
\usepackage{float} 
\usepackage{tabu}
\usepackage{textcomp}
\usepackage{mathtools}
\usepackage{hyperref}
\usepackage{amssymb}
\usepackage{bbm}
\usepackage{graphicx}
\usepackage{multicol}
\usepackage{multirow}
\usepackage[font=small,labelfont=bf]{caption}
\usepackage{subcaption}
\usepackage{algorithm}
\usepackage{afterpage}

\usepackage{bbm}
\usepackage{algorithmicx}
\let\labelindent\relax
\usepackage{enumitem,kantlipsum}
\usepackage{tablefootnote}
\usepackage{booktabs}
\usepackage{dsfont}
\usepackage[dvipsnames, svgnames]{xcolor} 
\usepackage{gensymb}
\usepackage{wrapfig}
\usepackage{capt-of}
\usepackage{xcolor}

\usepackage{algpseudocode}
\usepackage[backend=biber,
            hyperref=true,
            url=false,
            isbn=false,
            doi=false,
            backref=false,
            style=ieee,
            natbib=true,
            citestyle=numeric-comp,
            sorting=none,
            block=none]{biblatex}
\renewcommand{\bibfont}{\small}
\newcommand{\underlinedname}{\operatorname{\text{\textbf{\underline{R}}apid \textbf{\underline{A}}daptation of \textbf{\underline{P}}art\textbf{\underline{i}}cle \textbf{\underline{D}}ynamics}}}

\newcommand{\methodname}{\textsc{RAPiD}}
\newsavebox{\preliminaries}
\newcommand{\taskone}{\texttt{1D\_Inserting}}
\newcommand{\tasktwo}{\texttt{2D\_Covering}}
\let\ACMmaketitle=\maketitle
\renewcommand{\maketitle}{\begingroup\let\footnote=\thanks 
\ACMmaketitle\endgroup}
\makeatletter
\usepackage{graphicx}
\let\@fnsymbol\@arabic
\makeatother
\newcommand{\theappendix}{\@Alph\c@section}
\usepackage{etoolbox}
\usepackage{cuted}
\newcommand{\debugcounter}[1]{%
  \typeout{DEBUG: Counter #1 = \arabic{#1}}%
}

\pretocmd{\section}{\debugcounter{section}}{}{}
\pretocmd{\subsection}{\debugcounter{subsection}}{}{}
\allowdisplaybreaks
\title{\LARGE \bf Rapid Adaptation of Particle Dynamics
for Generalized Deformable Object Mobile Manipulation}

\author{Bohan Wu, Roberto Mart\'{i}n-Mart\'{i}n\textsuperscript{*}, and Li Fei-Fei\textsuperscript{*}
\thanks{* Equal Advising. Authors are with Stanford University and the University of Texas at Austin, USA. \texttt{\{bohanwu, feifeili\}@stanford.edu, robertomm@utexas.edu}}
}

\begin{document}
\maketitle
\begin{strip}
\includegraphics[width=\textwidth]{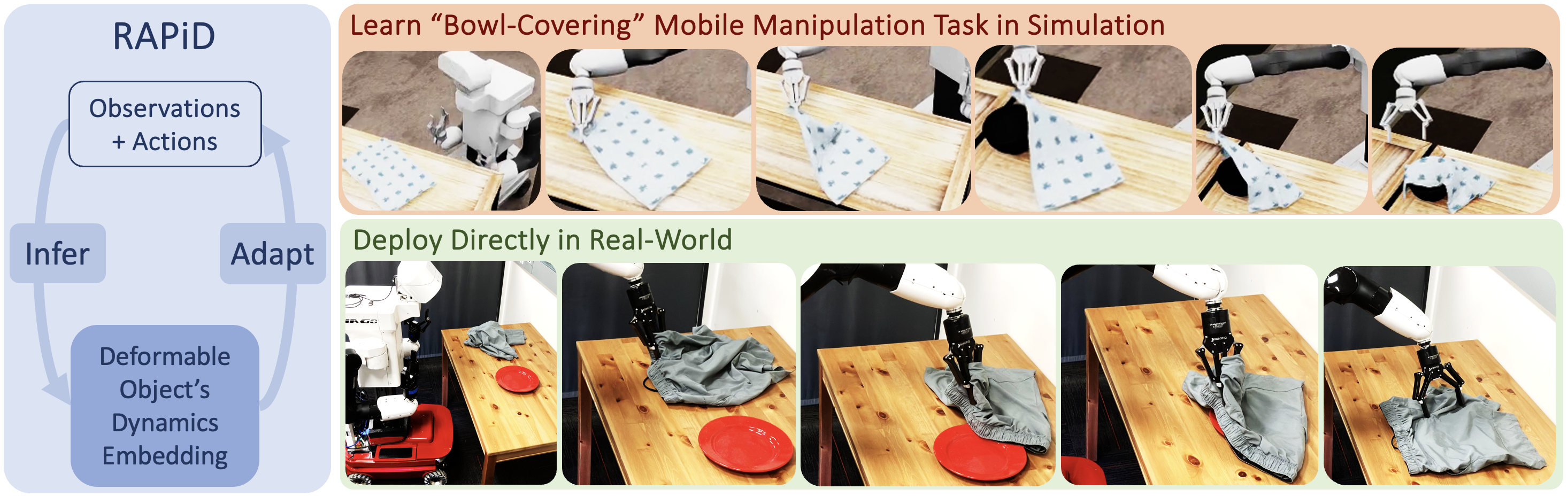}
\captionof{figure}{\small \textbf{\methodname{}: $\underlinedname$} (\textcolor{blue}{\textit{Left}}) is a method for learning to perform real-world deformable object mobile manipulation tasks by inferring and adapting to the unknown dynamics of deformable objects in real-time. \textcolor{Orange}{\textit{Top Right:}} In simulation, \methodname{} first learns a visuomotor policy to achieve a deformable object mobile manipulation task, such as bowl-covering, by inferring the deformable object's dynamics from the robot's visual observations and actions. \textcolor{OliveGreen}{\textit{Bottom Right}}: \methodname{} then deploys the learned visuomotor policy directly to the real world to achieve the deformable object mobile manipulation task, under \textbf{unseen object dynamics, instances, categories, and lighting conditions}, using only onboard sensor signals.}
\label{fig:pull}
\end{strip}

\begin{abstract}
We address the challenge of learning to manipulate deformable objects with unknown dynamics. 
In non-rigid objects, the dynamics parameters define how they react to interactions --how they stretch, bend, compress, and move-- and they are critical to determining the optimal actions to perform a manipulation task successfully.
In other robotic domains, such as legged locomotion and in-hand rigid object manipulation, state-of-the-art approaches can handle unknown dynamics using \textit{Rapid Motor Adaptation} (RMA). Through a supervised procedure in simulation that encodes each rigid object's dynamics, such as mass and position, these approaches learn a policy that conditions actions on a vector of latent dynamic parameters inferred from sequences of state-actions.
However, in deformable object manipulation, the object's dynamics not only include its mass and position, but also how the shape of the object changes.
Our key insight is that the recent ground-truth particle positions of a deformable object in simulation capture changes in the object's shape, making it possible to extend RMA to deformable object manipulation.
This key insight allows us to develop \methodname{}, a two-phase method that learns to perform real-robot deformable object mobile manipulation by: 1) learning a visuomotor policy conditioned on the object's dynamics embedding, which is encoded from the object's privileged information in simulation, and 2) learning to infer this embedding using non-privileged information instead, such as robot visual observations and actions, so that the learned policy can transfer to the real world. 
On a 22-DOF robot, \methodname{} enables 80\%+ success rates across two real-world vision-based deformable object mobile manipulation tasks, under unseen object dynamics, categories, and instances. More details are at \url{https://sites.google.com/view/rapid-robotics}.
\end{abstract}

\begin{refsection}[references.bib]
\section{Introduction}
\begin{figure*}[t]
\centering
\includegraphics[width=\textwidth]{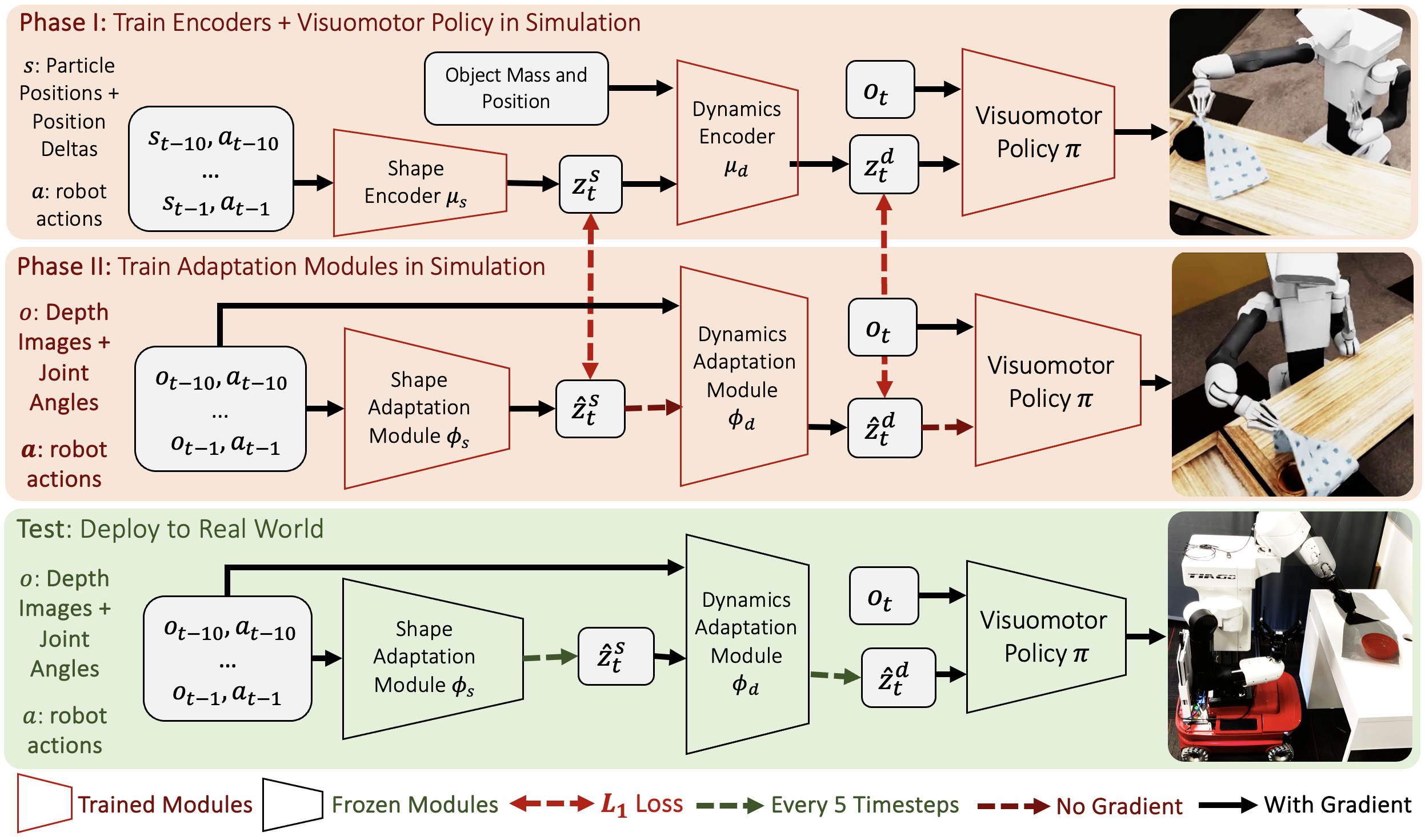}
\caption{\small \textbf{The \methodname{} Method}. 
Below we explain training (\textit{\textcolor{Orange}{top and middle}}) in simulation and deploying in the real world (\textit{\textcolor{OliveGreen}{bottom}}) a deformable object mobile manipulation solution with \methodname{}.
Training with \methodname{} (\textit{\textcolor{Orange}{top and middle}}) is a two-step procedure. In the first step (\textit{\textcolor{Orange}{top}}), the agent is asked to solve a simulated version of the task by learning a visuomotor policy and two encoders: a Shape Encoder, $\mu_{s}$, and a Dynamics Encoder, $\mu_{d}$, using privileged state information of the object from the simulator, such as mass and particle positions. In the second step (\textit{\textcolor{Orange}{middle}}), \methodname{} frees itself from privileged simulator information by learning a new Shape Adaptation module, $\phi_{s}$, and Dynamics Adaptation module $\phi_{d}$, using L1-losses to replace the Shape Encoder and the Dynamics Encoder, respectively. At test time (\textit{\textcolor{OliveGreen}{bottom}}), both the Shape and Dynamics adaptation modules and the visuomotor policy are directly deployed to the real world, with the adaptation modules updating the Shape Embedding $\hat{z}_t^{s}$ and Dynamics Embeddings $\hat{z}_t^{d}$ to the visuomotor policy once every 5 timesteps.}
\vspace{-10pt}
\label{fig:rda}
\end{figure*}
\label{s_intro}

Deformable objects, such as cables and clothes, can change in shape. As a result, the optimal strategies for deformable object manipulation often rely heavily on the object's dynamics, such as material properties, which are unknown a priori. For example, consider covering the top of a bowl with a 2D deformable object, such as a hat or a piece of cloth. For a hat, thanks to its high rigidity, a simple pick-and-place motion would be sufficient. But for a cloth, because of its softness, a fling followed by a sweep-and-cover motion is necessary. However, the dynamics of deformable objects are not fully observable a priori, and we as humans adapt to their dynamics in real-time \textit{during} our interaction with them. As we grasp and lift a hat or a cloth, we swiftly infer its material properties and adjust our placing motion or flinging speed to cover the bowl entirely. This rapid adaptation of unknown dynamics enables humans to efficiently manipulate deformable objects across diverse dynamics, categories, and instances, and we hope to endow robots with similar capabilities.

While many existing solutions can perform various deformable object manipulation tasks, generalizing these solutions to in-the-wild settings across diverse deformable object dynamics, categories, and instances remains challenging. For example, state estimation methods model deformable objects using state representations such as particles~\cite{jaillet1998deformable,li2018learning}, edges~\cite{triantafyllou2016geometric,yuba2017unfolding}, and graphs~\cite{longhini2023edo,lippi2020latent}, but they require full object observability and therefore cannot handle object occlusions. In parallel, a system identification method such as~\cite{chebotar2019closing} tries to estimate the deformable object's physics parameters from real-robot trajectories and then fine-tune the manipulation strategy in simulation, but this method requires multiple trajectories and therefore cannot adapt to unknown dynamics in real-time. Prior methods have also attempted to learn generalizable deformable object manipulation from large-scale cross-embodiment robotics datasets~\cite{black2024pi0,kim2024openvla}, but due to their limited scale, zero-shot generalization to a new robot embodiment remains challenging. Finally, other methods have also applied Rapid Motor Adaptation~\cite{kumar2021rma} to other robotics tasks such as legged locomotion~\cite{kumar2022adapting}, in-hand manipulation~\cite{qi2023hand}, and simulated robotic manipulation~\cite{liang2024rapid}. However, applying RMA to deformable object mobile manipulation presents an added complexity: the dynamics of the deformable object includes not only its mass and position, but also how the shape of the object changes. 

\begin{figure*}[t]
\centering
\includegraphics[trim=0cm 0cm 0cm 0cm, clip, width=\linewidth]{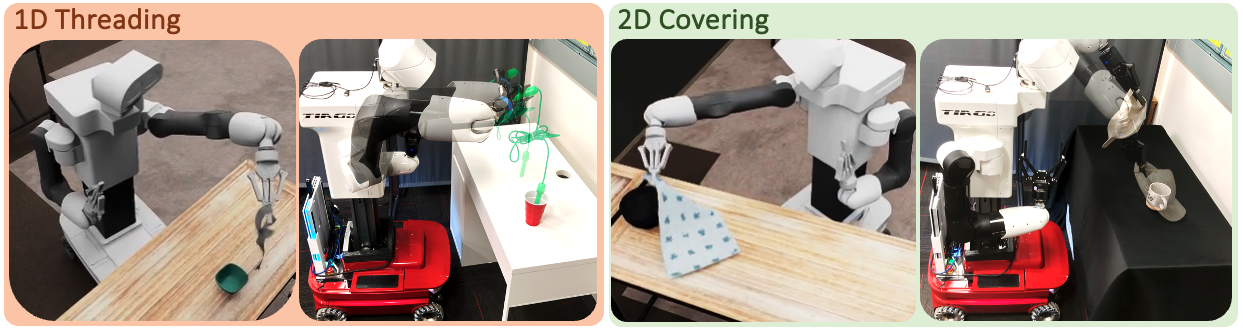}
\caption{\textbf{Simulated and Real-World Version of the Two Tasks}: \taskone~(\textcolor{Orange}{\textit{Left Half}}) and \tasktwo~(\textcolor{OliveGreen}{\textit{Right Half}}). Each task is trained in simulation (\textit{left image of each half}) and then deployed directly to the real world (\textit{right image of each half}) under unseen objects, environments, and lighting conditions. By interacting with deformable objects across a broad range of instances, geometries, and categories in simulation, \methodname{} successfully learns a visuomotor policy that adapts to and excels at manipulating real-world deformable objects of unseen dynamics, instances, and categories on a real bimanual mobile manipulator.}
\label{fig:tasks}
\vspace{-15px}
\end{figure*}
In this work, our key insight is that the recent ground-truth particle positions of a deformable object in simulation capture changes in the object's shape, making it possible to extend the RMA framework to learn to manipulate deformable objects with unknown dynamics.
This key insight allows us to create a novel two-phase method for learning to manipulate real-world deformable objects by inferring and adapting to their unknown dynamics: First, in a simulated version of the task, we train a visuomotor policy that is conditioned on a dynamics embedding of privileged simulator information, such as the recent ground-truth particle positions as well as the mass and position of the deformable object. 
Second, we learn a dynamics adaptation module in simulation to predict the dynamics embedding learned in the first step, using only non-privileged information, such as the robot's recent visual observations and actions. 
At test time, since the dynamics adaptation module and the visuomotor policy are learned in simulation using only onboard visual sensor signals, we can directly transfer them to a real robot to accomplish a deformable object mobile manipulation task. 
We call this method \textbf{\methodname{}} from \textbf{\underline{R}}apid \textbf{\underline{A}}daptation of \textbf{\underline{P}}art\underline{i}cle \textbf{\underline{D}}ynamics (Fig.~\ref{fig:pull}).

We evaluated \methodname{} in two real-robot deformable object mobile manipulation tasks involving 1D and 2D deformable objects with unseen dynamics, categories, and instances: \taskone~and \tasktwo. \methodname{} significantly outperforms existing state-of-the-art methods by 65\%+ success rates in both tasks based only on onboard visual inputs. This indicates that \methodname{}'s online adaptation algorithm provides a general and versatile mechanism for learning to manipulate deformable objects with unknown dynamics in simulation and then transferring the learned policies to the real world across a wide range of unseen deformable object dynamics, categories, and instances.
\section{Related Work}
\label{s_rw}
\textbf{Model-Based Deformable Objects Manipulation}. To learn deformable object manipulation with unknown dynamics, previous works have modeled deformable objects using various state representations, such as particle systems~\cite{jaillet1998deformable,li2018learning,liu2023robotic}, points~\cite{sun2014heuristic,weng2022fabricflownet,yuba2017unfolding,jangir2020dynamic}, graphs~\cite{triantafyllou2016geometric,yuba2017unfolding,longhini2023edo,lippi2020latent}, and differentiable simulators~\cite{li2018learning,chen2024differentiable,heiden2021disect,schenck2018spnets,xian2023fluidlab}. They then learn or plan manipulation strategies using demonstrations~\cite{chen2024differentiable,balaguer2011combining}, annotations~\cite{chi2024iterative}, or simulation~\cite{li2018learning,lin2022learning}. However, because state estimation can fail under hand-eye coordination or object occlusions~\cite{chen2024differentiable}, generalizing this representation to handle unseen object dynamics, categories, and instances in the real world remains challenging. In contrast, \methodname{} is a model-free method that learns to infer a dynamics embedding from recent robot visual observations and rapidly adapts to unknown deformable object dynamics in the real world without requiring state estimation. This allows \methodname{} to perform real-robot mobile manipulation tasks that require hand-eye coordination and robustness against object occlusions, such as the \taskone~and~\tasktwo~mobile manipulation tasks in Sec.~\ref{sec:experiments}.

\textbf{Rapid Motor Adaptation.} Our goal is to achieve rapid adaptation to unknown dynamics of deformable objects, therefore we revise previous works that apply Rapid Motor Adaptation~\cite{kumar2021rma} to other robotic tasks, such as quadruped locomotion~\cite{fu2022deep,fu2021coupling,fu2021minimizing,zhuang2023robot}, bipedal locomotion~\cite{kumar2022adapting,lee2020learning}, in-hand manipulation~\cite{qi2023hand}, and simulated robotic manipulation~\cite{liang2024rapid}. These methods first identify the task-relevant dynamics properties, such as mass and friction, obtain and encode the values of these properties from the simulator into a dynamics embedding, and learn a rapid adaptation module to infer this embedding from robot observations and actions. However, effective deformable object manipulation requires not only rapid inference and adaptation to the object's mass and position, but also how the shape of the objects changes, an element missing from these methods. To address this challenge, we revise \methodname{} to infer a dynamics 
embedding of both the deformable object's mass and position and the shape changes of the object, from its ground-truth physics parameters and recent particle positions in simulation, respectively, and then learns to infer this dynamics embedding from robot visual observations and actions instead. This allows \methodname{} to learn a visuomotor policy in simulation, transfer it to a real-robot mobile manipulator, and manipulate real-world deformable objects using only onboard sensors. 

\textbf{Learning Deformable Object Manipulation from Real-World Data.}
To learn generalizable deformable object manipulation, prior methods have collected real-world data such as random trajectories~\cite{wang2019learning}, human demonstrations~\cite{seita2019deep,seita2021learning,joshi2017robotic}, human annotations~\cite{lui2013tangled}, and large-scale teleoperation data~\cite{lee2015learning} for both policy pre-training and fine-tuning~\cite{black2024pi0,kim2024openvla,team2024octo,zhao2024aloha}. In particular, the last family first bootstraps learning from internet-scale vision-language pre-training and crowd-sourced cross-embodiment robotics datasets~\cite{o2024open}, before fine-tuning on robot-specific datasets to improve target-embodiment performance. While such large-scale real-world data collection methods have demonstrated impressive task complexity, they currently do not zero-shot generalize to new robot embodiments such as the TIAGo mobile manipulator used in our experiments, and whether it can collect enough data to achieve such zero-shot generalization remains to be seen. As a less labor-intensive alternative to this family of algorithms, \methodname{} learns deformable object mobile manipulation purely in simulation instead and zero-shot transfers to a vision-based mobile manipulator while generalizing across unseen object dynamics, categories, and instances.

\begin{figure*}[t]
\centering
\includegraphics[width=\linewidth]{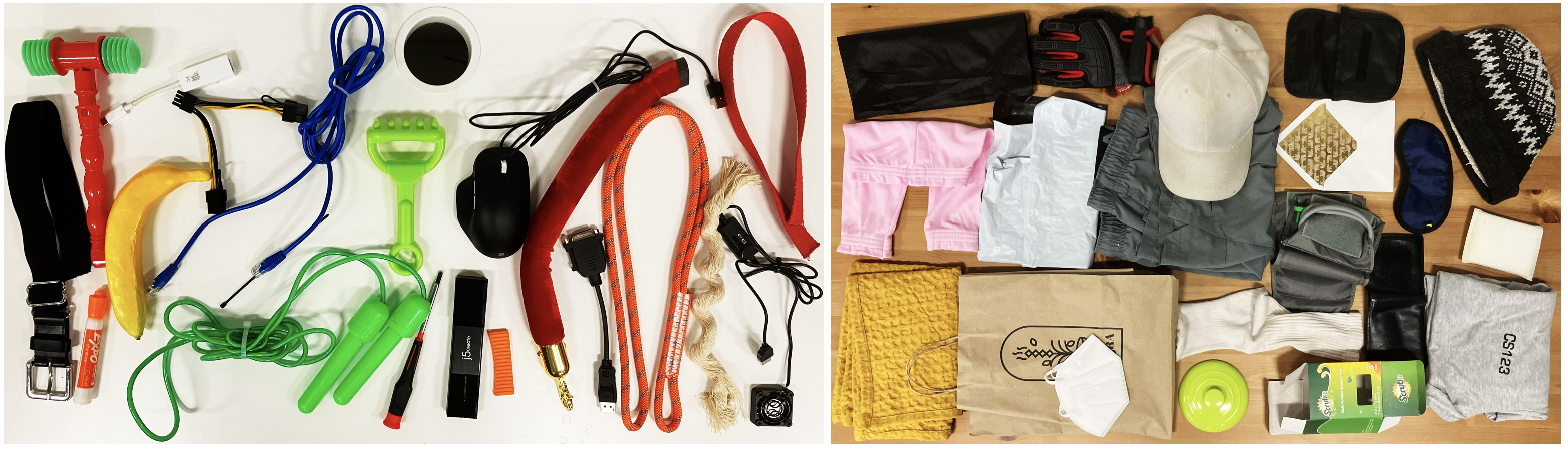}
\caption{All unseen real-world deformable objects used for the \taskone~(\textit{left}) and \tasktwo~(\textit{right}) mobile manipulation tasks, respectively. \textbf{1D categories (\textit{left}):} belt, hammer, marker, banana, HDMI adapter, GPU connector, jumping rope, ethernet cable, zip tie, rake, mouse, screwdriver, stick, rubber, velvet rope, VGA adapter, anchor rope, garden rope, ribbon, fan. \textbf{2D categories (\textit{right}):} polyester bag, pants, towel, plastic bag, face mask, paper bag, glove, shorts, cap, lid, sock, soft cuff, cardboard, envelope, pouch, wallet, eye-mask, t-shirt, sponge, wool hat. Learning from deformable objects across a wide range of instances in simulation enables \methodname{} to zero-shot transfer to real-world objects under unseen dynamics, instances, and categories.}
\label{fig:objects}
\vspace{-15px}
\end{figure*}
\textbf{Learning Deformable Object Manipulation from Simulation.}
To learn deformable object manipulation from simulation and transfer learned policies to the real world, previous works have adopted various strategies such as domain randomization~\cite{matas2018sim,hietala2021closing,ha2022flingbot,wu2019learning}, domain adaptation~\cite{chebotar2019closing}, real2sim~\cite{salhotra2022learning}, and dense object descriptors~\cite{sundaresan2020learning,ganapathi2021learning}. However, these methods do not include multiple visual observations and robot actions as input into their manipulation policy and therefore cannot infer and adapt to deformable object dynamics online. In particular, although Chebotar1 et al.~\cite{chebotar2019closing} proposes to adjust simulation parameters based on real-world experience, this method requires multiple real-world trajectories for adaptation training. It therefore cannot rapidly adapt to unknown dynamics in real-time. In comparison, \methodname{} leverages privileged simulator information to infer deformable object dynamics but then learns to infer them using robot visual observations and actions instead. This allows \methodname{} to rapidly adapt to a wide range of dynamics across real-world deformable objects, from soft objects (e.g., ribbons, cloths) to stiff items (e.g., cables, wallets).
\begin{figure*}[t]
\centering
\includegraphics[width=\linewidth]{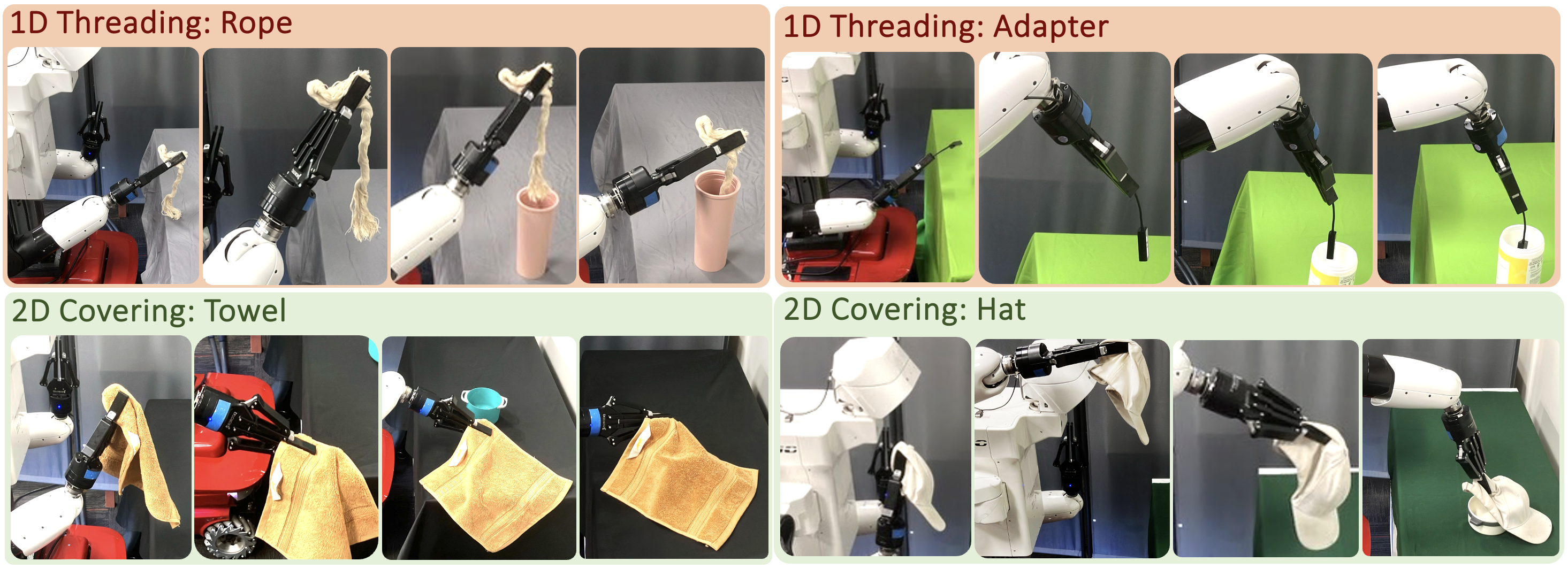}
\caption{\textbf{\methodname{}'s Dynamics Adaptive Behavior in} \taskone~(\textcolor{Orange}{\textit{Top}}) and \tasktwo~(\textcolor{OliveGreen}{\textit{Bottom}}) tasks. \textcolor{Orange}{\textit{Top Left:}} \methodname{} hangs the rope over the top of the container before lowering it. \textcolor{Orange}{\textit{Top Right:}} \methodname{} flips the gripper upside down before performing a direct adapter insertion. \textcolor{OliveGreen}{\textit{Bottom Left:}} \methodname{} moves the robot arm sideways and sweeps a towel horizontally from right to center to cover the bowl. \textcolor{OliveGreen}{\textit{Bottom Right}}: \methodname{} moves the arm vertically and places a hat directly on top of the bowl. This shows \methodname{}'s ability to infer different object dynamics and adapt actions accordingly to achieve the deformable object mobile manipulation tasks.}
\vspace{-15px}
\label{fig:rollout}
\end{figure*}
\section{\methodname{}: $\underlinedname$}
As shown in Fig.~\ref{fig:rda}, \methodname{} is a two-phase versatile method for learning to manipulate deformable objects in the real world by adapting to their unknown dynamics using raw visual inputs. In the first phase, \methodname{} learns a visuomotor mobile manipulation policy conditioned on two dynamics encoders using privileged ground-truth information obtained from simulation, such as ground-truth particle positions and position deltas of the deformable object  (Fig.~\ref{fig:rda}, \textit{top}). In the second phase, \methodname{} replaces the shape and dynamics encoders with two respective adaptation modules that only accept recent depth images and robot actions as input, making the entire algorithm free of privileged simulator information (Fig.~\ref{fig:rda}, \textit{middle}). At test time, \methodname{} freezes both the visuomotor policy and the two adaptation modules and deploys them directly to the real world (Fig.~\ref{fig:rda}, \textit{bottom}).

\methodname{} addresses both training phases as reinforcement learning (RL) problems which are modeled as Markov decision processes defined by the tuple $\mathcal{M} = \langle \mathcal{S}, \mathcal{O}, \mathcal{A}, \mathbb{T}, \mathcal{R}, \gamma, \rho_0, \mathcal{H}\rangle$. $\mathcal{S}$ is the state space (observable only in simulation). $\mathcal{O}$ is the space of observations. $\mathcal{A}$ is the action space. $\mathbb{T}$ is the dynamics model that governs the state transitions, $\mathbb{T}: \mathcal{S} \times \mathcal{A} \to \mathcal{S}$. $\mathcal{R}$ is a reward function. $\gamma \in [0, 1)$ is a discount factor. $\mathcal{H}$ is a finite horizon. $\rho_0$ is the distribution of initial states.
Reinforcement learning aims to train a policy, $\pi$, that takes optimal actions to maximize the expected cumulative future rewards across horizon $\mathcal{H}$.
Below, we describe each training and test phase in detail.

\textbf{Train Phase I: Learning a visuomotor policy in simulation conditioned on two ground-truth encoders}.
In the first phase, \methodname{} trains a visuomotor policy, $\pi$, that generates full-DOF robot actions to achieve the deformable object mobile manipulation task in simulation (Fig.~\ref{fig:rda}, \textit{top}). Concretely, the policy accepts a single depth image $o_t$ and a Dynamics Embedding $z_t^{d}$ as input. The Dynamics Embedding is encoded by a Dynamics Encoder $\mu_{d}$, which accepts the masses and positions of objects in simulation, and a Shape Embedding $z_t^{s}$ as input. As such, this Dynamics Encoder encodes dynamics information crucial to achieving the task. The Shape Embedding is, in turn, encoded by a Shape Encoder $\mu_{s}$, which accepts ground-truth object shape information as input, such as recent positions of the deformable object's particles and corresponding robot actions. As such, this Shape Encoder encodes information about how the shape of the object changes, which is crucial for achieving the task. We learn to encode dynamics and shape information in two separate encoders to quantify the dynamics vs. shape adaptation module's training losses we will encounter in the second phase. Because the dynamics and shape information are represented by low-dimensional privileged information from the simulator instead of high-dimensional visual observations, we train both the policy and the dynamics and shape encoders end-to-end using RL. 

Once \methodname{} has trained the visuomotor policy and the shape and dynamics encoders end-to-end, it learns two new adaptation modules --a shape adaptation module and a dynamics adaptation module--- to replace the shape and dynamics encoders, respectively, which we discuss next. 

\textbf{Train Phase II: Learning shape and dynamics adaptation modules in simulation}.
For \methodname{} to work in the real world, its policy, shape encoder, and dynamics encoder cannot accept any privileged ground-truth simulator information as input. Therefore, in this second training phase, the shape and dynamics encoders are replaced with their respective shape and dynamics adaptation modules that can be trained to infer the same dynamics embedding using only visual observations (Fig.~\ref{fig:rda}, \textit{middle}). Concretely, we first replace the Shape Encoder with a new Shape Adaptation module, $\phi_{s}$, which accepts recent depth images and robot joint angles and actions instead of ground-truth particle positions of the object as input. To train this Shape Adaptation module, we generate the ground-truth Shape Embedding, $z_t^{s}$, using the Shape Encoder and the inferred Shape Embedding, $\hat{z}_t^{s}$, from the Shape Adaptation module, and use L1-Loss to regress the inferred embedding to the ground-truth embedding. 
Similarly, we replace the Dynamics Encoder by training a new Dynamics Adaptation module, $\phi_{d}$, which accepts recent depth images and robot joint angles and actions instead of ground-truth dynamics information as input, using L1-loss between the ground-truth Dynamics Embedding $z_t^{d}$ and the inferred Dynamics Embedding $\hat{z}_t^{d}$. To prevent the Shape Adaptation module from encoding redundant dynamics information, we stop upstream gradients of training the Dynamics Adaptation module from flowing into the Shape Adaptation module. Finally, we fine-tune the visuomotor policy obtained during the first phase using the two new embeddings generated by the two new adaptation modules using RL. We stop upstream RL gradients from flowing into both adaptation modules so that the two modules separately encode the object dynamics and shape information-- even though they observe the same input-- that cannot be directly observed by the depth image $o_t$. The dynamics and shape encoders from Phase I are discarded before test time deployment, as discuss below.

\textbf{Test Time Deployment}. At test time, the robot is randomly placed in the real world with previously unseen deformable and rigid-body objects arranged randomly on a table. It needs to perform a task that requires using a deformable object with unknown dynamics. For each timestep, the robot interacts with objects in the environment using \methodname{}'s visuomotor policy, which accepts the current robot observation (i.e., a depth image and joint angles) and the Dynamics Embedding $\hat{z}_t^{d}$ as input and outputs full-DOF robot actions (Fig.~\ref{fig:rda}, \textit{bottom}). To generate the Dynamics Embedding, the robot stores a running buffer of the 10 most recent depth-image-observation-action pairs as input into both the shape and dynamics adaptation modules. To maintain short-term temporal behavioral consistency of the visuomotor policy, \methodname{} only periodically updates the shape and dynamics embeddings every 5 timesteps. This cycle repeats until the robot completes the task successfully.

\begin{figure*}[t]
\centering
\includegraphics[width=\linewidth]{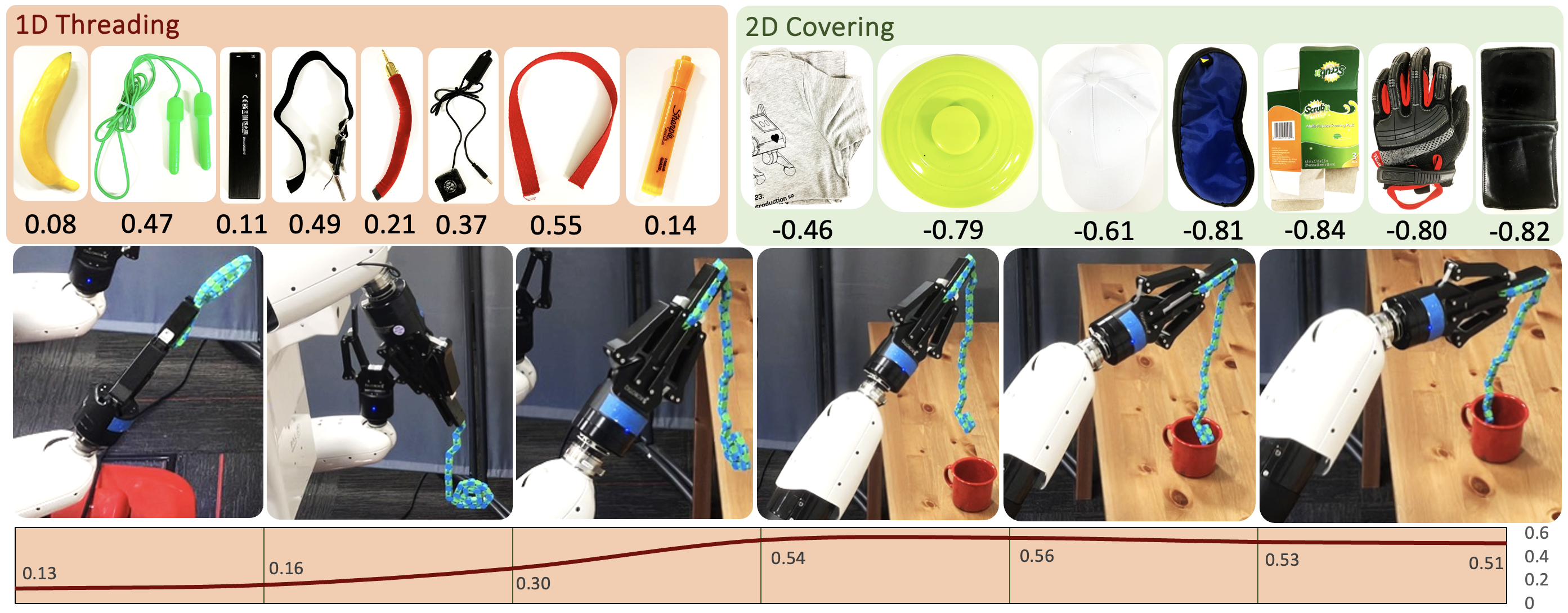}
\caption{\textbf{Visualization of \methodname{}'s dynamics embedding} in \taskone~(\textcolor{Orange}{\textit{Top-Left}}) and \tasktwo~(\textcolor{OliveGreen}{\textit{Top-Right}}) tasks. The number below each object photo is the value of a single dimension of the dynamics embedding. \textcolor{Orange}{\textit{Bottom}}: the rollout of the policy given a special object whose rigidity changes with rotation. The line charts below the rollout visualize the value of a single dimension of the dynamics embedding during robot execution. We see that certain dimensions of the dynamics embedding are strongly correlated with the softness of the objects, demonstrating the adaptation module's ability to infer the softness of the objects. }
\label{fig:embedding}
\end{figure*}
\section{Experiments}
\label{sec:experiments}
In our experiments, we use a TIAGo robot (Fig.~\ref{fig:tasks}), which is a 22-DOF bimanual mobile manipulator with 3-DOF for the omnidirectional base, 7-DOF for each of the two arms, 1-DOF torso, 2-DOF head, and 1-DOF for each of the two Robotiq parallel-jaw grippers. The robot has an RGB-D camera mounted on its head. 
We use depth images of $224 \times 224$ resolution captured at 3 Hz. 
In both simulation and the real world, \methodname{}'s input is the robot's depth image and joint angles.

We use OmniGibson~\cite{li2023behavior, li2024behavior} as the training simulator for \methodname{}. In OmniGibson, A model of the same TIAGo robot is placed in front of a simulated version of the task (Fig.~\ref{fig:tasks}). To train the visuomotor mobile manipulation policy, we use a binary reward of 1 for task success and a small reward negatively correlated with the distance between the deformable object and the target rigid-body object on the table.
In the real world, the robot is placed in front of an unseen table with unseen deformable and rigid-body objects (Fig.~\ref{fig:tasks}). 
\methodname{} then controls the robot's bimanual mobile manipulation capabilities (base, torso, head, grippers, and arms) to grasp and manipulate the deformable object to achieve the task, using all DOFs simultaneously.
All experiments are conducted autonomously without human intervention from raw visual inputs with no QR codes. 
We use real-world objects (Fig.~\ref{fig:objects}), environments (Fig.~\ref{fig:tasks}), and lighting conditions (Fig.~\ref{fig:tasks}) never seen during simulation training. 
We repeat each task 20 times with different initial locations and objects. 

\textbf{Tasks}: We evaluate \methodname{} across two challenging deformable-object mobile manipulation tasks that require online adaptation to the deformable object's unknown dynamics a priori: \taskone~and \tasktwo. In \taskone~(Fig.~\ref{fig:tasks}, \textit{left}), the goal is to pick up a 1D deformable object on the table and insert either one of its two ends into a container. In the real-world version of the task, we place a 1D deformable object randomly on the table, selected across 20 categories such as rope, cables, and belts~(Fig.~\ref{fig:objects}, \textit{left}), and a container selected across 20 object categories such as cups, bowls, and pans. Success is defined by whether the robot inserts one end of the deformable object into the container within 300 seconds. 
In \tasktwo~(Fig.~\ref{fig:tasks}, \textit{right}), the goal is to pick up a 2D deformable object on the table and cover the opening of a container on the table-- a representative task in household scenarios. In the real-world version of the task, we place a 2D deformable object randomly on the table, sampled across 20 categories such as lids, napkins, towels, and plastic films~(Fig.~\ref{fig:objects}, \textit{right}). There is no parameter randomization applied to object pose, material stiffness, or friction, since such randomization is achieved implicitly via object instance randomization. Success is defined by whether the robot successfully covers 90\% of the container's opening within 300 seconds. 
Achieving both tasks requires closed-loop hand-eye coordination, robustness against object deformation and occlusions, and online adaptation to the unknown dynamics of the deformable object during real-time interaction, as shown in Fig.~\ref{fig:rollout} and \ref{fig:embedding}.

\begin{table*}[t]
\centering
\small
\caption{\textbf{Successful Trials} Out of 20 and Success Rates for \methodname{}, Baselines and Ablations}
\label{tab:results}
\resizebox{\linewidth}{!}{
\begin{tabular}{|c|c|c|c|c|c|c|c|}
\toprule
 & \methodname{} & DMfD & DDOD & \methodname{}-No-Adapt & a\methodname{}-No-Shape & \methodname{}-E2E \\
\midrule
\taskone & 17 (85\%) & 3 (15\%) & 2 (10\%) & 7 (35\%) & 7 (35\%) & 5 (25\%) \\ 
\tasktwo & 16 (80\%) & 1 (5\%)  & 5 (25\%) & 5 (25\%) & 9 (45\%) & 4 (20\%) \\\midrule
Total & 33 / 40 (82.5\%) & 4 / 40 (10\%) & 7 / 40 (17.5\%) & 12 / 40 (30\%) & 16 / 40 (40\%) &  9 / 40 (22.5\%)\\
\bottomrule
\end{tabular}}
\end{table*}
\textbf{Baselines}: We compare \methodname{} to two baselines for learning deformable object manipulation: DMfD~\cite{salhotra2022learning} (Table~\ref{tab:results}, \textit{column 3}) and DDOD~\cite{sundaresan2020learning,ganapathi2021learning} (Table~\ref{tab:results}, \textit{column 4}). The first baseline, DMfD, learns deformable object stationary manipulation of 1D and 2D object categories from simulated expert demonstrations. The second baseline, DDOD, learns dense object descriptor representations for 1D~\cite{sundaresan2020learning} and 2D~\cite{ganapathi2021learning} deformable objects and learns deformable object stationary manipulation from a single human demonstration using this representation. To make DDOD comparable to \methodname{}, which requires no demonstrations, we script a policy based on the dense object descriptors learned from DDOD to perform deformable object manipulation, instead of providing human demonstrations. To help both baselines overcome failures in grasping the deformable object and acquire navigation capabilities comparable to \methodname{}, we use \methodname{} to execute the grasping and navigation of the tasks before activating the baseline for the deformable object manipulation part of the tasks. Learning deformable object mobile manipulation with unseen deformable object dynamics, categories, and instances using real-robot imitation learning~\cite{black2024pi0,kim2024openvla,team2024octo,zhao2024aloha,zhao2023learning,nair2017combining,brohan2023rt,brohan2022rt,liu2024rdt} requires collecting an infeasible number of real-world demonstrations on the TIAGo robot, which makes them incomparable to \methodname{}, since \methodname{} requires no real-robot demonstrations. Extending \methodname{} to fine-tune on real-world data to compare to these methods is left for future work.

We also conduct three ablation experiments of \methodname{}: 1) \textit{\methodname{}-\textit{No-Adapt}} (Table~\ref{tab:results}, \textit{column 5}), which removes both the Shape and Dynamics Adaptation Modules and trains \methodname{}'s visuomotor policy only; 2) \textit{a\methodname{}-No-Shape} (Table~\ref{tab:results}, \textit{column 6}), which deletes the Shape Adaptation Module and trains only the visuomotor policy and the Dynamics Adaptation module; and 3) \textit{\methodname{}-E2E} (Table~\ref{tab:results}, \textit{column 7}), which ignores both Shape and Dynamics Encoders, skips the first of the two training phases and trains \methodname{} with end-to-end RL without the two L1 losses.

In our experiments, we answer four main questions:

\textbf{Q1: How does \methodname{} compare to state-of-the-art sim2real methods in real-robot deformable object mobile manipulation?} In Table~\ref{tab:results} (columns 2-4), \methodname{} outperforms both baselines significantly across both tasks. Qualitatively, DMfD fails in both \taskone~and \tasktwo~because after grasping, the deformable object suffers from severe deformation and occlusions by the robot gripper during the hand-eye-coordinated motions of inserting and covering, preventing DMfD from generating a sufficiently accurate segmentation mask for the deformable object. In some cases where the segmentation mask is accurate enough, DMfD is still unable to adapt to the unknown dynamics of the deformable object. DDOD fails in \taskone~and \tasktwo~because the deformable object suffers from in-the-wild, non-top-down camera viewpoints, severe deformation, and robot gripper occlusions, making DDOD incapable of predicting accurate dense object descriptors for the deformable object during manipulation.
In some cases where the dense object descriptor predictor performs well, DDOD is still unable to adapt to the unknown dynamics of the deformable object. In contrast, \methodname{} performs closed-loop 6-DOF control and hand-eye coordination directly from non-top-down depth images and outperforms both DMfD and DDOD. 

\textbf{Q2: How important are \methodname{}'s Shape and Dynamics Adaptation Modules to deformable object mobile manipulation?} We compare \methodname{} (Table~\ref{tab:results}, \textit{column 2}) to \textit{\methodname{}-\textit{No-Adapt}} (Table~\ref{tab:results}, \textit{column 5}), which ablates Shape and Dynamics Adaptation Modules. We observe a 52.5\% drop in performance for \methodname{}-\textit{No-Adapt}. Qualitatively, in both \taskone~and \tasktwo, \methodname{}-\textit{No-Adapt} treats deformable objects as rigid objects, which fails when the object is made of soft materials. Therefore, \methodname{}'s Adaptation Modules are crucial.

\textbf{Q3: How important is \methodname{}'s ability to infer the object's shape changes?} We compare \methodname{} to \textit{a\methodname{}-No-Shape} (Table~\ref{tab:results}, \textit{column 6}), which removes the Shape Adaptation Module. We observe that \methodname{} encounters a 42.5\% success rate degradation compared to our full variant using both the Shape and Dynamics Adaptation Modules. In both \taskone~and \tasktwo, we observe that \textit{a\methodname{}-No-Shape} is successful at grasping the deformable object and moving it closer to the target container to insert or cover, but unable to execute the precise dynamic motion needed to insert or cover the container, which requires inferring the deformable object's shape changes. Thus, the ability to infer how the object's shape changes is crucial. 

\textbf{Q4: Is optimizing both adaptation modules to mimic their corresponding ground-truth encoders necessary, or is end-to-end RL sufficient for task success?} We compare \methodname{} to \textit{\methodname{}-E2E} (Table~\ref{tab:results}, \textit{column 7}), which trains the entire architecture with end-to-end RL. We observe that \textit{\methodname{}-E2E} fails to converge optimally on its RL policy gradients during training and experiences a 60\% success rate reduction: due to the high-dimensionality of the 10 most recent depth image observations and unstable RL training gradient signals, \methodname{} relies solely on current depth observations to perform the task, ignoring and failing to learn a useful dynamics embedding. Thus, optimizing both adaptation modules using L1-loss in a two-phase training process is crucial. 
\section{Conclusion}
We proposed \methodname{}, a method for learning to perform real-world deformable object mobile manipulation tasks by inferring and adapting to the unknown dynamics of deformable objects in real-time. \methodname{} results from a two-phase training procedure that first infers a deformable object dynamics embedding from the object's recent ground-truth particle positions, and then infers this same embedding using recent robot visual observations and actions instead. \methodname{} achieved success in two deformable object mobile manipulation tasks across unknown deformable object dynamics, categories, and instances. 

\renewcommand*{\bibfont}{\footnotesize}
\printbibliography
\newpage
\end{refsection}

\end{document}